\documentclass[twocolumn]{IEEEtran}
\renewcommand{\baselinestretch}{1.0}

\ifCLASSINFOpdf
\else
\usepackage[dvips]{graphicx}
\fi
\usepackage{url}

\usepackage{graphicx}
\usepackage{cite}
\usepackage{times}
\usepackage{epsfig}
\usepackage{amsmath}
\usepackage{amssymb}
\usepackage{epstopdf}
\usepackage{subfig}
\usepackage{algorithm}
\usepackage{algorithmic}
\usepackage{threeparttable}

\usepackage{hyperref}

\usepackage{makecell}

\usepackage{amsfonts}

\usepackage{booktabs}
\usepackage{multirow}

\usepackage{caption} 

\begin{document}

\title{Semantically Informed Salient Regions Guided Radiology Report Generation}

\author{Zeyi Hou, Zeqiang Wei, Ruixin Yan, Ning Lang, Xiuzhuang Zhou, \emph{Member, IEEE }
\thanks{This work is partially supported by the National Natural Science Foundation of China under grants 61972046, the Beijing Natural Science Foundation under grant Z190020, and the Proof of Concept Program of Zhongguancun Science City and Peking University Third Hospital under grant HDCXZHKC2022202.
(\emph{Corresponding author: Xiuzhuang Zhou})

Zeyi Hou, Zeqiang Wei, and Xiuzhuang Zhou are with the School of Artificial Intelligence, Beijing University of Posts and Telecommunications, Beijing, China. (xiuzhuang.zhou@bupt.edu.cn).

Ruixin Yan and Ning Lang are with the Department of Radiology, Peking University Third Hospital, Beijing, China. (langning800129@bjmu.edu.cn)
}
}

\markboth{Journal of \LaTeX\ Class Files, Vol. xx, No. xx, July 2024}
{Shell \MakeLowercase{\textit{et al.}}: Bare Demo of IEEEtran.cls for IEEE Journals}
\maketitle

\begin{abstract}
  Recent advances in automated radiology report generation from chest X-rays using deep learning algorithms have the potential to significantly reduce the arduous workload of radiologists.
  However, due to the inherent massive data bias in radiology images, where abnormalities are typically subtle and sparsely distributed, existing methods often produce fluent yet medically inaccurate reports, limiting their applicability in clinical practice.
  To address this issue effectively, we propose a Semantically Informed Salient Regions-guided (SISRNet) report generation method.
  Specifically, our approach explicitly identifies salient regions with medically critical characteristics using fine-grained cross-modal semantics.
  Then, SISRNet systematically focus on these high-information regions during both image modeling and report generation,
  effectively capturing subtle abnormal findings, mitigating the negative impact of data bias, and ultimately generating clinically accurate reports.
  Compared to its peers, SISRNet demonstrates superior performance on widely used IU-Xray and MIMIC-CXR datasets.
\end{abstract}


\IEEEpeerreviewmaketitle

\section{Introduction}
\IEEEPARstart{C}{hest} radiography is currently the most widely used medical imaging examination, playing a crucial role in clinical diagnosis \cite{alyasseri2022review} and epidemiological studies \cite{rahman2021exploring}. 
Crafting detailed and accurate radiology reports is a complex and time-consuming task.
Consequently, deep learning models for the automated interpretation of X-rays have garnered significant attention, as these models have the potential to greatly reduce radiologists' workload and enhance clinical efficiency.

Automated Radiology Report Generation (RRG) essentially involves transforming intricate visual data from chest X-rays into comprehensive textual descriptions.
Unlike traditional image captioning \cite{vinyals2015show,plummer2015flickr30k}, RRG presents a highly challenging inter-modal translation task for several reasons:
Firstly, radiological data exhibit significant bias, with normal instances dominating the dataset, while abnormal regions typically occupy only a small portion of the image in pathological cases, making it challenging for deep learning models to effectively capture abnormal features.
Moreover, RRG generates detailed and long reports comprising multiple sentences to accurately and adequately describe medical observations within a X-ray image. 
These sentences exhibit complex semantic relationships with corresponding image regions.

Existing RRG works primarily adopt an encoder-decoder architecture to convert visiual input into textual output,
where a fixed-length semantic vector serves as the bridge between the two components.
However, for the complex cross-modal task of generating a diagnostic report from a chest X-ray image,
this vector struggles to fully capture all the information from the input image, especially given the challenges of medical data bias.
As a result, it limits the decoder's ability to generate satisfactory reports.
To address these challenges, several studies have enhanced the encoder-decoder framework by 
introducing attention mechanism\cite{song2022cross,gajbhiye2022translating,gu2023automatic}, reinforcement learning\cite{miura2021improving,delbrouck2022improving}, knowledege graphs\cite{li2023dynamic,yan2023attributed,wang2022embracing} or other auxilliary tasks\cite{liu2019clinically,nguyen2021automated,hou2023organ,yang2023radiology,liu2024multi}.
While these approaches have proven effective in RRG, they remain inadequate for capturing exceedingly sparse and subtle lesions in chest X-ray images with consistent anatomical structures.

Recently, some methods have been developed to alleviate the negative impact of medical data bias by focusing on key areas in X-ray images.
These approaches leveraged class activation mapping (CAM)\cite{wang2024camanet}, anatomical region detection\cite{basu2024focusmae}
or medical keywords guidance\cite{xie2024rethinking} to identify key regions in chest X-ray images, ensuring focused feature extraction from these regions.
These methods help reduce interference from irrelevant background regions in understanding medical semantics. 
However, they rely on costly annotations, lack contextual information and are too coarse to accurately identify subtle lesions.
Additionally, they fail to capture critical details such as the presence, location, and severity of lesions.
More importantly, while these methods enhance the general representation of X-ray images, they offer limited support for generating long reports with complex and scattered semantics.

\begin{figure*}[t]
   \setlength{\abovecaptionskip}{3pt}  
   \setlength{\belowcaptionskip}{0pt}   
   \centering
   \includegraphics[width=\textwidth]{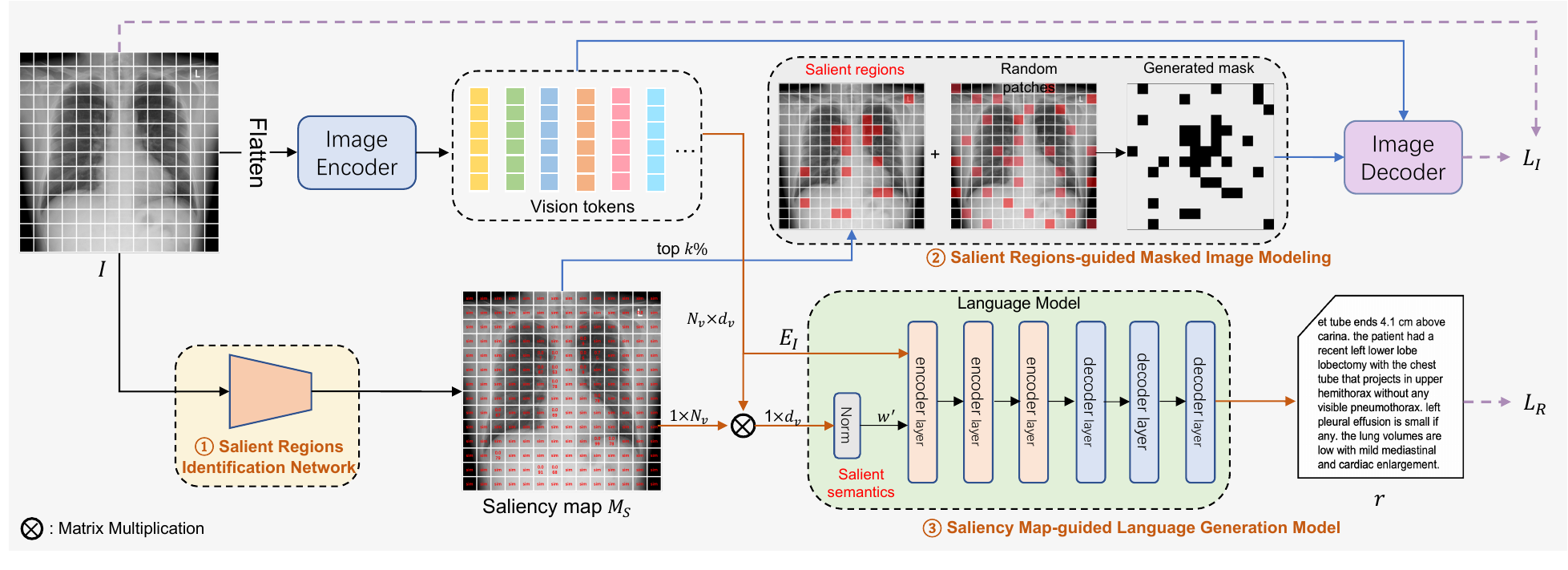}
   \captionsetup{justification=centering, font=small} 
   \caption{Overview of the proposed radiology report generation method.}
   \label{fig:Overview}
   \end{figure*}

Medical images exhibit consistent anatomical structures while displaying subtle individual variations.
These variations typically appear as pathological lesions that alter the morphology and texture of tissues or organs,
and we refer to them as \emph{salient regions}.
These salient regions are usually sparse and visually inconspicuous, but play a vital role in clinical diagnosis.
In this letter, we propose a novel Semantically Informed Salient Regions-guided (SISRNet) report generation method to generate clinically accurate reports.
Specifically, we employ fine-grained cross-modal alignment to extract medical semantics from radiology reports to identify semantically informed salient regions that contain pathological clues.
Then, SISRNet prioritizes these high-information regions in both image modeling and report generation, effectively capturing subtle abnormalities in chest X-ray images to generate clinically accurate reports.
This work makes the following contributions:

\noindent\textbullet\ We effectively identify semantically informed salient regions rich in critical insights, addressing the limitations of previous methods that detect coarse regions and overlook key contextual cues.

\noindent\textbullet\ These semantically informed salient regions are leveraged to guide both image modeling and report generation to generate clinically accurate reports.

\noindent\textbullet\ Comprehensive experiments and ablation studies on widely used IU-Xray dataset and MIMIC-CXR dataset demonstrate the effectiveness of our method.

\section{Related work}

\subsection{Encoder-Decoder Architecture}
Most report generation models adopt an encoder-decoder architecture, where the encoder extracts features from the input X-ray image, and the decoder utilizes these features to generate the report.
The encoder-decoder framework was originally designed for machine translation\cite{cho2014learning,sutskever2014sequence}. 
Its success in this field inspired researchers to apply it to the task of natural image caption and medical report generation, and became the cornerstone of almost all subsequent methods for generating chest X-ray reports. 
Recently, several studies\cite{chen2021cross,shang2022matnet,zhang2023semi,zhang2024visual} have proposed improved networks and modules to enhance the encoder-decoder framework for generating high-quality diagnostic reports.
For instance, R2GenCMN\cite{chen2021cross} introduced shared memory to improve cross-modal interaction by recording relationships between image and text. 
Similarly, MATNet\cite{shang2022matnet} and VCIN\cite{zhang2024visual} designed refined encoder and decoder networks to enhance report quality.
Despite these advancements, the fixed-length semantic representation in the encoder-decoder architecture poses challenges for complex cross-modal generation tasks, such as generating comprehensive reports from chest X-ray images.
This limitation restricts the model's ability to fully capture and express all relevant semantics, leading to incomplete textual descriptions. 

To address this issue, subsequent studies have integrated techniques such as attention mechanism\cite{song2022cross,gajbhiye2022translating,gu2023automatic}, reinforcement learning\cite{miura2021improving,delbrouck2022improving}, knowledge graphs\cite{li2023dynamic,yan2023attributed,wang2022embracing}, or other auxilliary tasks\cite{liu2019clinically,nguyen2021automated,hou2023organ,yang2023radiology,liu2024multi} into the encoder-decoder framework to improve the quality of RRG.
For example, Song\cite{song2022cross} proposed a cross-modal contrastive attention model to effectively obtain pathological features, which compares the input features extracted by the encoder with the most similar instance features retrieved from the database.
Miura \cite{miura2021improving} introduced two types of reinforcement rewards: Exact Entity Match Reward and Entailing Entity Match Reward, which are designed to enhance the agent's performance in generating diagnostic reports.
DCL\cite{li2023dynamic} proposed a knowledge graph with dynamic structure and nodes to facilitate diagnostic report generation.
In addition, studies \cite{liu2019clinically,nguyen2021automated,hou2023organ} employed an image classification task to predict disease labels and assist the main task of report generation.
Although these methods have been shown to be effective in RRG, they are still insufficient to capture extremely sparse and subtle lesions in chest X-ray images with consistent anatomical structures.
\begin{figure*}[t]
   \setlength{\abovecaptionskip}{3pt}  
   \setlength{\belowcaptionskip}{0pt}   
   \centering
   \includegraphics[width=\textwidth]{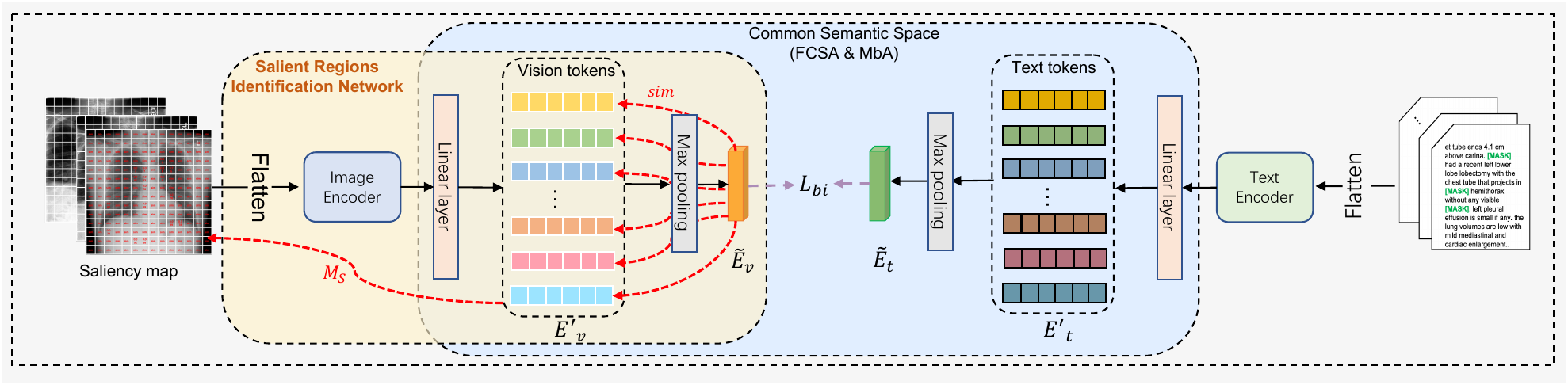}
   \captionsetup{justification=centering, font=small} 
   \caption{Salient regions identification network.}
   \label{fig:SISR}
   \end{figure*}
\subsection{Key Regions-based Report Generation}
Similar to our approach, several previous methods have been proposed to mitigate the negative impact of medical data bias by emphasizing critical regions.
For example, CAMANet \cite{wang2024camanet} introduced class activation map (CAM) to guide report generation and use the attention mechanism to make the model pay more attention to the class activation area.
However, CAM lacks contextual information and cannot accurately reflect medical semantics.
Some studies \cite{huang2023mgmae,bandara2023adamae,basu2024focusmae,xie2023medim,xie2024rethinking} explored masking strategies for Masked Autoencoders (MAE) to enhance medical image representation.
FocusMAE \cite{basu2024focusmae} employed anatomical regions detection to select masking areas.
This method requires costly annotations and still struggles to detect abnormalities, as the anatomical regions are too coarse to accurately identify subtle lesions.
MedIM \cite{xie2024rethinking} leveraged keywords and random sentence in report to determine masking patches.
However, this strategy lacks contextual understanding and fail to capture critical details in images, such as the presence, location, and severity of lesions.
More importantly, while these methods improve the general representation of X-rays, they provide limited assistance for generating long reports with complex and scattered semantics.
In contrast, our method identifies semantically informed salient regions and piroritizes these high-information regions in both image modeling and report generation,
detecting subtle abnormalities and generating clinically accurate reports.

\section{Approach}
The overall architecture of our proposed SISRNet for report generation is illustrated in Fig.~\ref{fig:Overview}, consisting of the following components:
A salient regions identification network identifies high-information regions in X-ray images;
A salient regions-guided masked image modeling captures subtle abnormalities;
A saliency map-guided language generation model preserves sparse symptom clues in reports.
In particular, the salient regions identification network operates independently from the other components.
We first train the identification network and freeze its parameters, then use it to identify salient regions for other components.

\subsection{Semantically Informed Salient Regions Identification}
Medical image-text pairs naturally available in radiograph datasets,
and we argue that fine-grained cross-modal alignment enables the learning of discriminative clues in reports, such as the presence, location and severity of lesions.
To identify salient regions associated with these discriminative clues, we construct a corss-modal common semantic space, where a \emph{saliency map} is created to reflect the importance of each image patch.
As shown in Fig.~\ref{fig:SISR}, the common semantic space is constructed using Fine-grained Cross-modal Semantics Alignment (FCSA) and Mapping before Aggregate (MbA) \cite{wei2023masked} strategy.
In our approach, The application of MbA in identifying salient regions not only minimizes the missing of critical information but also ensures the comparability between local and global features (both in the common semantic space).
Compared to X-ray image feature learning approaches guided by anatomical regions or class activation mapping, the proposed method leverages fine-grained textual semantics to extract more refined salient regions with implicit contextual clues.

The process of identifying salient regions begins with extracting fine-grained features from the images and the reports.
For image patch feature extraction, we use the Vision Transformer (ViT) ViT-B16 \cite{dosovitskiy2020image} as the vision encoder,
where an image $I\in\mathbb{R}^{H\times W\times C}$ is devided into $N_v$ non-overlapping patches
to obtain vision token features $E_v\in\mathbb{R}^{N_v \times d_v}$, where $d_v$ is the encoding dimension.
Similarly, for token feature extraction, BioClinicalBERT \cite{alsentzer2019publicly} is employed to tokenize the report $r$ to $N_t$ subword tokens and extract text token features $E_t\in \mathbb{R}^{ N_t  \times d_t }$.

The visual token features $E_v$ and text token features $E_t$ are initially flattened and linearly projected into local visual embeddings $E_v^{\prime}=\left\{e^v_1,e^v_2,\cdots,e^v_{N_v}\right\}$ and local textual embeddings $E_t^{\prime}=\left\{e^t_1,e^t_2,\cdots,e^t_{N_t}\right\}$.
The local features of both modalities are then aggregated into thier respective global features using max pooling:

\begin{equation} \label{equ:MbA}
   \begin{split}
  \tilde{E}_v &= MaxPool \left( E_v^{\prime}  \right) ,  E_v^{\prime} = f^{\prime}_v \left( E_v \right)\\
  \tilde{E}_t &= MaxPool \left( E_t^{\prime} \right)   , E_t^{\prime} = f^{\prime}_t \left( E_t \right)
   \end{split}
\end{equation}
where $f^{\prime}_v$ and $f^{\prime}_t$ denote the linear layer for vision token features and text token features respectively,
and $MaxPool$ is the maximum pooling operation.

This mapping before aggregation pattern preserves the sparse semantic distribution of chest X-ray images and reports, laying a solid foundation for subsequent fine-grained cross-modal alignment.
We employ a bidirectional and asymmetrical cross-modal contrastive loss $L_{bi}$ to learn the common semantic space:
\begin{equation} \label{equ:Contrastive_loss}
   \begin{split}
   L_{v \to t}    &= -\frac{1}{B} \sum_{i=1}^{B} \log_{}{\frac{\exp \left( sim \left( \tilde{E}_{v,i} , \tilde{E}_{t,i} \right) / \tau \right)}{\sum_{j=1}^{B} \exp \left( sim \left( \tilde{E}_{v,i}, \tilde{E}_{t,j}\right) / \tau \right)} }  ,    \\
   L_{t \to v}    &= -\frac{1}{B} \sum_{i=1}^{B} \log_{}{\frac{\exp \left( sim \left( \tilde{E}_{t,i} , \tilde{E}_{v,i} \right) / \tau \right)}{\sum_{j=1}^{B} \exp \left( sim \left( \tilde{E}_{t,i}, \tilde{E}_{v,j}\right) / \tau \right)} }   ,  \\
   L_{bi}= & \lambda_{v} L_{v \to t} + \lambda_{t} L_{t \to v}  ,
   \end{split}
\end{equation}
where $\lambda_{v}$ and $\lambda_{t}$ are weight parameters of different directions.
$B$ denotes the batch size. $sim \left( \cdot,\cdot \right)$ represents the cosine similarity between two vectors and $\tau$ is the temperature parameter.
Intuitively, the loss from image modality to text modality $L_{v \to t}$ pulls the representation of the $i$-th image sample closer to its corresponding text sample while pushing it away from other text samples.
Similarly, the loss from text modality to image modality $L_{t \to v}$ brings the $i$-th text sample closer to its corresponding image sample while distancing it from other image samples.
By optimizing contrastive loss in both directions simultaneously, the representation spaces of the image and text modalities are effectively aligned.

After cross-modal contrastive learning, both global image features $\tilde{E}_v$ and local visual features $\left\{e^v_1,e^v_2,\cdots,e^v_{N_v}\right\}$ in the common semantic space imply the corresponding textual semantic information.
The collaboration of MbA and max pooling enables cosine similarity
\begin{equation} \label{equ:Sim}
sim \left( \left\{e^v_1,e^v_2,\cdots,e^v_{N_v}\right\} , \tilde{E}_v \right)
\end{equation}
to represent the contribution of image patches to discriminative semantics and constitutes a \emph{saliency map} $M_S$.
Since the aggregation operation is performed in the learned common semantic space, as shown in Fig.~\ref{fig:SISR},
the vision tokens whose features are most similar (top $k\%$) to the global features can be considered as \emph{salient regions} that containing critical semantics,
such as lesions, pathological abnormalities, postoperative changes, and implanted medical devices, \textit{etc.}

\subsection{Salient Regions-guided Image Modeling}
In the clinical workflow, radiologists typically focus first on abnormal regions before composing the corresponding discriptions.
Motivated by this, we propose a salient regions-guided masked image modeling to enhance the representation of subtle abnormalities in X-ray images.
The image is first input into a pre-trained salient regions identification network to obtain salient regions $S$, which containing rich pathological semantics, as described above.
Masking salient regions when training MAE can helps the visual encoder effectively learn more refined representations of abnormalities, thereby mitigating the negative impact of medical data bias.
In addition, we also mask patches scattered throughout the entire image to enhance the model's ability to represent all anatomical regions.
The masking probability for $i$-th vision token $x^v_i$ is:
\begin{equation} \label{equ:Mask_probability}
   p_i \leftarrow p_i + I \left( x^v_i \in S \right) \varphi  ,  \  p_i \sim U(0,1),
\end{equation}
where $I\left( \cdot \right)$ is indicator function.
Intuitively, if an image patch is identified as a semantically informed salient region, 
\textit{i.e.}, $x^v_i \in S$, its mask probability $p_i$ increases by $\varphi$.
The initial mask probability follows a uniform distribution $U(0,1)$,
enabling random masking across the entire chest X-ray image. 
The masking strategy described in Eqn.(\ref{equ:Mask_probability}) allows the model to effectively capture subtle lesion details while preserving the overall characteristics of the chest X-ray.

An image decoder is used for reconstructing the pixels of masked patches with the MSE loss:
\begin{equation} \label{equ:RRG_MAE}
   L_{I} = \frac{1}{ \left | M \right | } \sum_{i \in M}  \left \| z_i - \hat{z}_i   \right \|_2  ,
\end{equation}
where $\hat{z}_i$ and $z_i$ represent the output by the image decoder and the ground-truth patch, respectively.
$M$ is the number of masked patches.
The salient regions-guided masked image modeling encourages the encoder of RRG to explore refined medical image representations.

\subsection{Saliency Map-guided Report Generation}
Similar to chest X-ray images, radiology reports are dominated by normal descriptions, and the sentences in a report typically convey distinct meanings.
The decoder of RRG model struggles to capture and describe various individual abnormalities in different sentences.
Consequently, semantically informed guidance is also crucial during the decoding stage to generate clinically accurate reports.
We propose to integrate the saliency map into the generation process of the language model to enrich the discriminative information.
Specifically, the discriminative representation $w^{\prime} \in \mathbb{R}^{ 1  \times d_v }$ is acquired by matrix multiplication
between the saliency map $M_S \in \mathbb{R}^{ 1  \times N_v }$ and the image patch features $E_I\in\mathbb{R}^{N_v \times d_v}$ followed by a normalization layer:
\begin{equation} \label{equ:Discriminative_Representation}
   w^{\prime} = Norm(M_S E_I)  ,
\end{equation}
where $w^{\prime}$ captures the pathological clues of each region.
We add $w^{\prime}$ as an extra input token into the language model, whose self-attention mechanism gradually incorporates the saliency map for text representation.

The report generation loss is defined as a cross entropy of the ground truth words $y$ and predicted words $\hat{y}$,
\begin{equation} \label{equ:RRG_CE}
   L_{R} = - \frac{1}{l} \sum_{i=1}^{l}\sum_{j=1}^{V} y_{ij} \log_{}{\hat{y}_{ij}}   ,
\end{equation}
where $l$ and $V$ denote the document length and the vocabulary size respectively.
$\hat{y}_{ij}$ represents the confidence of selecting the $j$-th word in the vocabulary for the $i$-th position in the generated report.

\subsection{Training Process}
The salient regions identification network in Fig.~\ref{fig:Overview} is trained independently from the other components of SISRNet, ensuring separate gradient flows.
Specifically, SISRNet first trains the salient regions identification network with the loss:
\begin{equation} \label{equ:loss_1}
   L_1 = \lambda_v L_v + \lambda_t L_t + \lambda_{bi} L_{bi} ,
\end{equation}
where $\lambda_v$ ,$\lambda_t$ and $\lambda_{bi}$ denote the weight parameters of image reconstruction loss $L_v$, text completion loss $L_t$ and bidirectional cross-modal contrastive loss $L_{bi}$, respectively.

Once trained, the parameters of the salient regions identification network are frozen and used to detect semantically informed salient regions in chest X-ray images.
After identifying the salient regions, the masked image modeling network and the language generation network are trained using the loss:
\begin{equation} \label{equ:loss_2}
   L_2 = \lambda_I L_I + \lambda_R L_R ,
\end{equation}
where $\lambda_I$ and $\lambda_R$ represent the weight parameters of $L_I$ and $L_R$.

%
\begin{table*}[t!]
   \begin{center}
   \captionsetup{justification=centering, font=footnotesize} 
   \caption{\MakeUppercase{Performance comparison of different methods on Natural Language Generation (NLG) metrics and Clinical Efficacy (CE) metrics on the IU-Xray and MIMIC-CXR datasets.}}
   \label{table:Metrics}
   \renewcommand{\arraystretch}{1.1}
   \setlength{\tabcolsep}{3.6pt}{
   \begin{tabular}{l|clcc|cccccccc|cccc}
       \toprule\noalign{\smallskip}
      \multirow{2}*{Dataset}        & &\multirow{2}*{Method}                       & &\multirow{2}*{Year}  &  &              \multicolumn{6}{c}{NLG Metrics}                            &  &  & \multicolumn{3}{c}{Micro CE Metrics}           \\  \noalign{\vspace{1pt}}  \cline{7-12} \cline{15-17} \noalign{\vspace{1pt}} 
                                    & &                                            & &                     &  & BLEU-1       & BLEU-2      & BLEU-3   & BLEU-4   & METEOR   & ROUGE-L   &  &  & Precision  & Recall  & F1-score        \\ 
       \noalign{\vspace{1pt}} \cline{1-17}\noalign{\vspace{3pt}}  
       \multirow{11}*{IU-Xray}       
                                    & &R2GenCMN \cite{chen2021cross}               & & 2021                &  &0.475         &0.309        &0.222     &0.170     &0.191     &0.375      &  &  & -          & -       & -                \\
                                    & &XPRONet \cite{wang2022cross}                & & 2022                &  &0.525    &0.357        &0.262     &0.199     &0.220     &0.411      &  &  & -          & -       & -                \\
                                    & &KiUT \cite{huang2023kiut}                   & & 2023                &  &0.525    &0.360        &0.251     &0.185     &0.242     &0.409      &  &  & -          & -       & -                \\
                                    & &DCL \cite{li2023dynamic}                    & & 2023                &  &N/A             &N/A            &N/A         &0.163     &0.193     &0.383      &  &  & -          & -       & -                \\
                                    & &ATAG \cite{yan2023attributed}               & & 2023                &  &N/A             &N/A            &N/A         &N/A       &N/A          &0.341      &  &  & -          & -       & -                \\
                                    & &M2KT \cite{yang2023radiology}               & & 2023                &  &0.497         &0.319        &0.230      &0.174    &N/A         &0.399      &  &  & -          & -       & -                \\
                                    & &OaD \cite{li2024organ}                      & & 2024                &  &0.514         &0.347        &0.253     &0.193     &0.213     &0.385      &  &  & -          &-        & -        \\
                                    & &SILC \cite{liu2024multi}                    & & 2024                &  &0.472         &0.321        &0.234     &0.175     &0.192     &0.379      &  &  & -          &-        & -        \\
                                    & &PhraseAug \cite{mei2024phraseaug}           & & 2024                &  &0.510         &0.369        &0.285     &0.231     &0.218     &0.431      &  &  & -          &-        & -        \\
                                    & &CAMANet \cite{wang2024camanet}              & & 2024                &  &0.504         &0.363        &0.279     &0.218     &0.203     &0.404      &  &  & -          & -       & -                \\
      \noalign{\vspace{1pt}} \cline{2-17}\noalign{\vspace{3pt}}                               
                                    & &SISRNet                                      & & ours               &  &{\bf0.529}    &{\bf0.380}   &{\bf0.290}   &{\bf0.234}  &{\bf0.256}     &{\bf0.436}      &  &  & -          & -       & -                 \\
       \noalign{\vspace{1pt}}\hline\noalign{\vspace{1pt}}
       \multirow{11}*{MIMIC-CXR}
                                    & &R2GenCMN \cite{chen2021cross}               & & 2021                &  &0.353         &0.218        &0.148     &0.106     &0.142     &0.278      &  &  &0.334        &0.275         &0.278       \\
                                    & &XPRONet \cite{wang2022cross}                & & 2022                &  &0.344         &0.215        &0.146     &0.105     &0.138     &0.279      &  &  &0.463        &0.285         &0.353       \\
                                    & &KiUT \cite{huang2023kiut}                   & & 2023                &  &0.393         &0.243        &0.159     &0.113     &0.160    &0.285      &  &  &0.371        &0.318         &0.321      \\
                                    & &DCL \cite{li2023dynamic}                    & & 2023                &  &N/A             &N/A            &N/A         &0.109     &0.150     &0.284      &  &  &0.471        &0.352         &0.373       \\
                                    & &ATAG \cite{yan2023attributed}               & & 2023                &  &N/A            &N/A            &N/A         &N/A         &N/A         &0.227      &  &  &N/A          &N/A           & 0.395                \\
                                    & &M2KT \cite{yang2023radiology}               & & 2023                &  &0.386         &0.237        &0.157     &0.111     &N/A         &0.274      &  &  & 0.420          & 0.339           & 0.352             \\
                                    & &OaD \cite{li2024organ}                      & & 2024                &  &0.412         &0.262        &0.178     &0.127     &0.156     &0.284      &  &  &0.364        &0.382         &0.372       \\
                                    & &SILC \cite{liu2024multi}                    & & 2024                &  &0.406         &0.267        &0.190     &0.141     &0.163     &0.309      &  &  &0.457        &0.337         &0.330    \\
                                    & &PhraseAug \cite{mei2024phraseaug}           & & 2024                &  &0.447         &0.314        &0.235     &0.184     &0.208     &0.353      &  &  &N/A            &N/A            &0.391    \\
                                    & &CAMANet \cite{wang2024camanet}              & & 2024                &  &0.374         &0.230        &0.155     &0.112     &0.145     &0.279      &  &  &0.483        &0.323         &0.378       \\
      \noalign{\vspace{1pt}} \cline{2-17}\noalign{\vspace{3pt}}  
                                    & &SISRNet                                    & & ours           &  &{\bf0.471}      &{\bf0.347}    &{\bf0.274}  &{\bf0.209}  &{\bf0.211}  &{\bf0.367}  &  &  &{\bf0.551}   &{\bf0.467}    &{\bf0.509}   \\
       \bottomrule
   \end{tabular}  
   }
   \end{center}
   \end{table*}

\section{Experiments}

\subsection{Datasets and Experimental Settings}
We conduct a range of experiments on radiology report generation using two widely-used public datesets, \textit{i.e.}, IU-Xray \cite{demner2016preparing} and MIMIC-CXR \cite{johnson2019mimic}.
IU-Xray contains 7,470 images and 3,955 corresponding reports. For a fair comparison, we follow the same data split
as \cite{wang2024camanet,chen2020generating,chen2021cross} to divide the IU-Xray dataset into 70\%/10\%/20\% for train/validation/test set.
MIMIC-CXR is the largest radiology dataset with 227,835 medical reports, associated with 377,110 X-ray images.
The official data split is adopted for the MIMIC-CXR dataset.

The most widely used metrics in RRG, \emph{i.e.}, Natural Language Generation (NLG) metrics and Clinical Efficacy (CE) metrics, are employed to evaluate model performance.
The NLG metrics (BLEU-1 to BLEU-4 \cite{papineni2002bleu}, METEOR \cite{banerjee2005meteor}, and ROUGE-L \cite{lin2004rouge}) assess language generation performance by counting matching $n$-grams.
However, they are ill-suited for capturing the diagnostic accuracy of generated reports \cite{boag2020baselines,liu2019clinically}.
Following \cite{chen2020generating,tanida2023interactive}, we also report the CE metrics to measure the clinical correctness of generated reports.
The CE metrics compare the presence of 14 clinical observations extracted by CheXbert \cite{smit2020combining} between generated and reference reports.
Due to the imbalance among different clinical observation labels in the MIMIC-CXR dataset, we calculate the CE scores by micro-averaging across the 14 observations.

We resize images to (256, 256) and then cropped to (224, 224).
The ViT extract 14 × 14 visual tokens.
The language model used to generate reports is a randomly initialized Transformer with 3 layers, 8 attention heads, and 512 dimensions hidden states. 
We set the masking rate to 75\% in report generation network.
The vision tokens corresponding to the top 20\% of local features are selected as salient regions, and the probability increment $\varphi$ is set to 0.35.
During training, we set the weight parameters as follows: $\lambda_{v \to t}=0.75$, $\lambda_{t \to v}=0.25$, $\lambda_v=1.0$, $\lambda_t=1.0$, $\lambda_{bi}=0.1$, $\lambda_I=1.0$ and $\lambda_R=0.1$.

\subsection{Quantitative and Qualitative Studies}
We compare our method with several state-of-the-art models for RRG, including R2GenCMN \cite{chen2021cross}, XPRONet \cite{wang2022cross}, KiUT \cite{huang2023kiut},
DCL \cite{li2023dynamic}, ATAG \cite{yan2023attributed}, M2KT \cite{yang2023radiology}, OaD \cite{li2024organ}, SILC \cite{liu2024multi}, PhraseAug \cite{mei2024phraseaug}, and CAMANet \cite{wang2024camanet}.
These methods are selected because they share the most similar experimental settings with our approach, such as the dataset split and the beam search size.
The performance of different RRG methods on IU-Xray and MIMIC-CXR test sets is shown in Table~\ref{table:Metrics}.
Our approach outperforms the alternatives on NLG metrics across two datasets.
This demonstrates that the proposed method can generate high-quality diagnostic reports.
Due to the diverse word orders and sentence structures in english reports, evaluating sentence similarity solely by counting matching $n$-grams provides only a preliminary assessment of text quality.
For practical clinical applications, the diagnostic accuracy of the generated reports is a more crucial metric.
The proposed SISRNet achieves superior performance across all three CE metrics, significantly improving the precision and recall of chest disease diagnosis.
This demonstrates that the salient regions-guided report generation effectively captures subtle abnormal findings and mitigates the negative impact of massive data bias in the medical domain, generating clinically accurate reports.

In recent years, the field of large language models (LLMs) has rapidly advanced, leading to the emergence of numerous medical LLMs designed to tackle various image-to-text tasks \cite{wu2023towards,li2024llava,moor2023med,chen2024internvl,he2024meddr}.
To comprehensively evaluate the proposed SISRNet method, we compare it against several state-of-the-art medical LLMs on MIMIC-CXR dataset.
The compared models include RadFM\cite{wu2023towards}, LLaVA-Med\cite{li2024llava}, Med-Flamingo\cite{moor2023med}, InternVL\cite{chen2024internvl} and MedDr\cite{he2024meddr}.
We adopt the evaluation metrics proposed in MedDr\cite{he2024meddr} and the results of different methods in generating chest X-ray diagnostic reports are presented in Table~\ref{table:Metrics_LLM}.
The proposed SISRNet demonstrates significant advantages over medical LLMs.
It achieves substantially higher scores than existing medical LLMs in BLEU-1, BLEU-4, and ROUGE-L metrics, while achieving comparable results with the state-of-the-art model MedDr\cite{he2024meddr} in the METEOR metric.
This result highlights the effectiveness of semantically informed salient regions in enhancing the quality of generated reports.
Moreover, these findings further indicate that although medical large language models exhibits strong generalization capabilities and can handle a wide range of medical image-to-text tasks, their performance on specific tasks often lags behind that of specialized models.
Considering that large language models typically require larger datasets, more advanced hardware, and longer training and inference times, they cannot fully replace task-specific models in the practical application of automated chest X-ray report generation.

\begin{table}[t]
   \begin{center}
   \renewcommand{\arraystretch}{1.1}
   \captionsetup{justification=centering, font=footnotesize} 
   \caption{\MakeUppercase{Performance comparision with different medical LLMs on the MIMIC-CXR dataset.}}
   \label{table:Metrics_LLM}
   \setlength{\tabcolsep}{6pt}{
   \begin{tabular}{lcccc}
   \toprule\noalign{\smallskip}
   Methods & BLEU-1 & BLEU-4 & METEOR & ROUGE-L  \\
   \noalign{\smallskip}
   \hline
   \noalign{\smallskip}
   RadFM\cite{wu2023towards}                   & {0.221}                 & {0.056}                & {0.204}                & {0.205}     \\
   LLaVA-Med\cite{li2024llava}                 & {0.193}                 & {0.010}                & {0.132}                & {0.139}       \\
   Med-Flamingo\cite{moor2023med}              & {0.224}                 & {0.019}                & {0.140}                & {0.145}     \\
   InternVL\cite{chen2024internvl}             & {0.255}                 & {0.017}                & {0.175}                & {0.156}      \\
   MedDr\cite{he2024meddr}                     & {\underline{0.322}}     & {\underline{0.072}}    & {\bf0.238}             & {\underline{0.226} }     \\
   \noalign{\vspace{1pt}} \cline{1-5}\noalign{\vspace{3pt}} 
   SISRNet                                     & {\bf0.471}              & \bf{0.209}             & {\underline{0.211}}    & {\bf0.367}      \\
   \bottomrule
   \end{tabular}  }
   \end{center}
\end{table}

\begin{table}[t]
   \captionsetup{justification=centering, font=footnotesize} 
   \begin{center}
   \caption{\MakeUppercase{Ablation results on the components of SISRNet on the MIMIC-CXR dataset.}}
   \label{table:Ablation-results}
   \renewcommand{\arraystretch}{1.1}
   \setlength{\tabcolsep}{10pt}{
   \begin{tabular}{l|ccc}
   \toprule\noalign{\smallskip}
   Method                                 & Precision    & Recall     & F1-score   \\
   \noalign{\vspace{1pt}} \cline{1-4}\noalign{\vspace{3pt}} 
   BASE                                   &0.420        &0.227         &0.294        \\
   BASE + SISR-Masking                    &0.542        &0.463         &0.499        \\
   BASE + SISR-LM                         &0.476        &0.318         &0.380        \\
   \noalign{\vspace{1pt}} \cline{1-4}\noalign{\vspace{3pt}} 
   SISRNet                              &{\bf0.551}   &{\bf0.467}    &{\bf0.509}  \\
   \bottomrule
   \end{tabular}  }
   \end{center}
   \end{table}

\textit{Ablation Studies:} 
We conduct a series of ablation experiments on MIMIC-CXR, the largest and most widely used dataset, to evaluate the negative impact of different components of SISRNet and the probability of masking semantically informed salient regions on the quality of the generated reports. 
We use the Micro CE metric, which includes precision, recall, and F1-score, to assess the quality of reports generated under different settings.

The ablation study results for different components of SISRNet are presented in Table~\ref{table:Ablation-results}. 
The "BASE" model represents the encoder-decoder model with random masked image modeling and original language model.
"BASE + SISR-Masking" replaces the random masking strategy in "BASE" model with a masking strategy guided by semantically informed salient regions.
"BASE + SISR-LM" replaces the original language model in "BASE" model with a report generation model guided by the saliency map.
We observe that the "BASE" model yields poor results because it fails to capture subtle abnormalities in X-rays.
Each component of SISRNet enhances the clinical accuracy of the generated reports, and the systematic combining of "SISR-Masking" and "SISR-LM" achieves superior overall results, generating clinically accurate reports.

\begin{table}[t]
   \renewcommand{\arraystretch}{1.1}
   \begin{center}
   \captionsetup{justification=centering, font=footnotesize} 
   \caption{\MakeUppercase{Ablation results on masking probability of salient regions on the MIMIC-CXR dataset.}}
   \label{table:Ablation_results_varphi}
   \setlength{\tabcolsep}{10pt}{
   \begin{tabular}{l|ccc}
   \toprule\noalign{\smallskip}
   Methods                                 & Precision    & Recall     & F1-score  \\
   \noalign{\vspace{1pt}} \cline{1-4}\noalign{\vspace{3pt}} 
   Random Masking                          &0.420        &0.227         &0.294        \\
   \noalign{\vspace{1pt}} \cline{1-4}\noalign{\vspace{3pt}} 
   SISRNet($\varphi=0.25$)                  &0.525        &0.443         &0.466       \\
   SISRNet($\varphi=0.3$)                 &0.536        &0.423         &0.472       \\
   SISRNet($\varphi=0.35$)                  &{\bf0.551}   &{\bf0.467}    &{\bf0.509}  \\
   SISRNet($\varphi=0.4$)                  &0.529        &0.430         &0.474       \\
   SISRNet($\varphi=0.5$)                  &0.511        &0.391         &0.443       \\
   \bottomrule
   \end{tabular}  }
   \end{center}
   \end{table}
\begin{figure}[h]
   \centering
   \includegraphics[width=8.6cm]{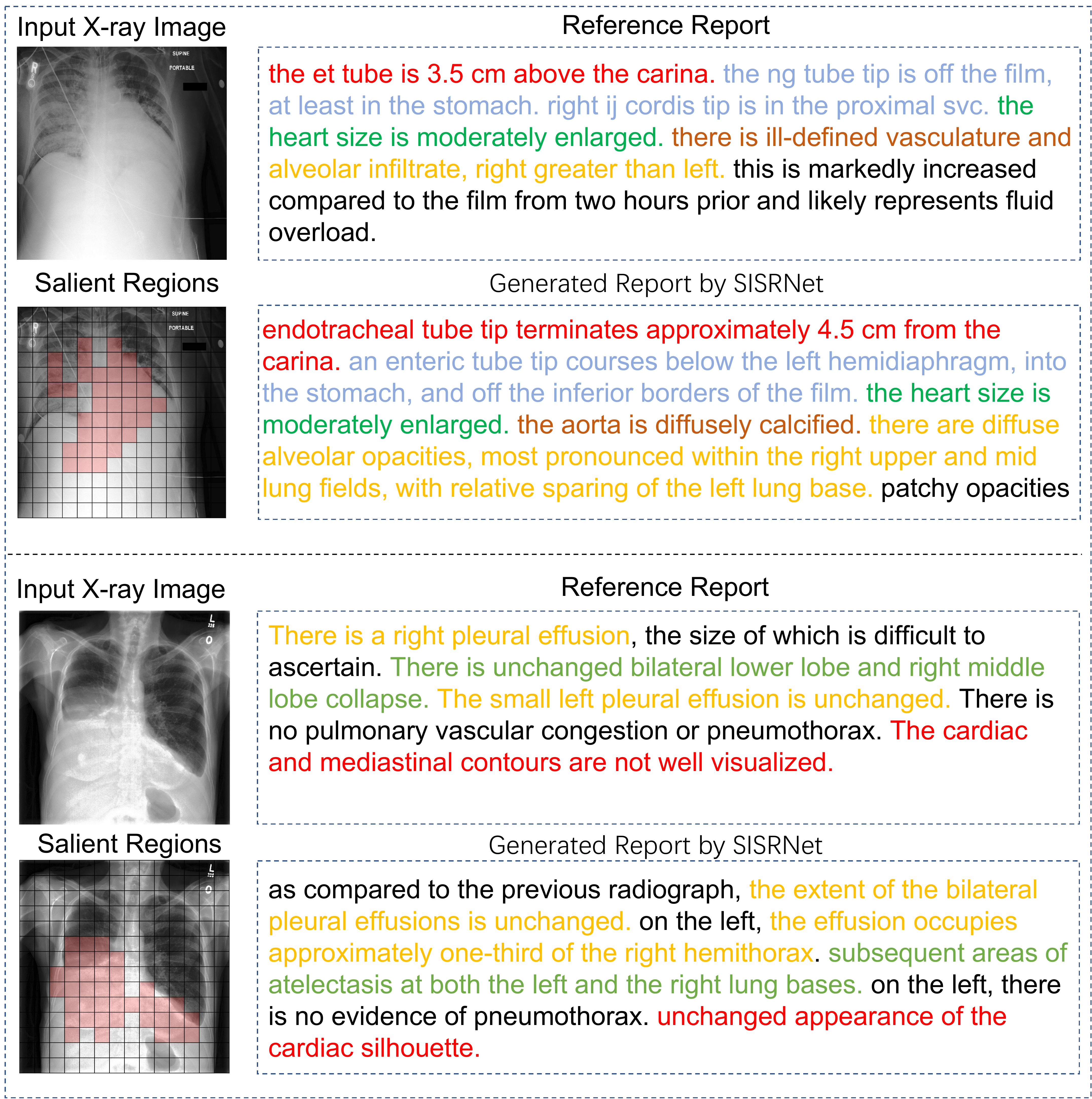}
   \captionsetup{justification=centering, font=small} 
   \caption{Case of reports generated by SISRNet on the MIMIC-CXR dataset. (Consistent color highlights similar clinical observations.)}
   \label{fig:Visualization}
   \end{figure}
The ablation results on the masking probability of salient regions in SISRNet are presented in Table~\ref{table:Ablation_results_varphi}.
The results show that using a random masking strategy leads to suboptimal reports, with lower precision, recall, and F1-scores compared to other settings.
This is because random masking assigns the same masking probability to all image patches, making it difficult for the model to capture fine-grained, discriminative abnormalities in chest X-ray images.
Increasing the masking probability for salient regions to a certain extent improves the quality of generate reports.
However, excessive masking probability degrades model performance by affecting feature learning in the reconstruction process.
If the masking probability increment is too small, the variation in masking probability is negligible, making the semantic guidance of salient regions ineffective.
Conversely, if the increment is too large, most salient region image patches are completely masked, hindering feature extraction in image reconstruction.
To balance these factors, we constrain the masking probability increment $\varphi$ within [0.25, 0.5] in our ablation experiments.
As shown in Table~\ref{table:Ablation_results_varphi}, the optimal diagnostic report quality is achieved when $\varphi$ is set around 0.35 in experiments on the MIMIC-CXR dataset.

\textit{Case Study:} Examples of salient regions and reports generated by SISRNet are presented in Fig.~\ref{fig:Visualization}
We observe that the salient regions effectively highlight pathological areas such as endotracheal tube, cardiomegaly, atelectasis, pleural effusion and lung opacity in the X-ray image.
It also shows that SISRNet improves the interpretability of the model.
The generated report accurately describes medical observations, including abnormalities, location, approximate size, and other relevant details.

\section{Concluesion}
We have introduced a novel Semantically Informed Salient Regions-guided (SISRNet) report generation method
designed to mitigate the negative impact of significant data bias and generate medically accurate reports.
SISRNet identifies salient regions with abundant information using fine-grained cross-modal semantics,
and then systematically prioritizes these high-information regions in both image modeling and report generation.
Experimental results demonstrate that the proposed method generates fluent reports that accurately describe the clinical findings.

{\small
\bibliographystyle{ieeetr}
\bibliography{myreference}
}


\end{document}